\documentclass[preprints,article,accept,moreauthors,pdftex]{mdpi}
\usepackage[caption=false]{subfig}
\usepackage{float}
\usepackage{tipa}

\firstpage{1}
\pubvolume{1}
\issuenum{1}
\articlenumber{0}
\pubyear{2021}
\copyrightyear{2020}
\datereceived{}
\dateaccepted{}
\datepublished{}
\hreflink{https://doi.org/} 

\Title{Title}

\Title{Deep Neural Network Based Respiratory Pathology Classification Using Cough Sounds}

\TitleCitation{Title}

\Author{Balamurali B T $^{1}$, Hwan Ing Hee $^{2}$, Saumitra Kapoor $^{1}$,  Oon Hoe Teoh  $^{3}$, Sung Shin Teng $^{4}$, Khai Pin Lee $^{4}$, Dorien Herremans $^{5}$ and Jer Ming Chen $^{1}$}

\AuthorCitation{Lastname, F.; Lastname, F.; Lastname, F.}

\address{%
$^{1}$ \quad
Science, Mathematics and Technology, Singapore University of Technology and Design, Singapore 487372, Singapore; balamurali\_bt@sutd.edu.sg\\
$^{2}$ \quad Department of Paediatric Anaesthesia, KK Women’s and Children’s Hospital, Singapore 229899, Singapore; e-mail@e-mail.com\\
$^{3}$ \quad Respiratory Medicine Service, Department of  Paediatrics, KK Women’s and Children’s Hospital, Singapore 229899, Singapore; e-mail@e-mail.com\\
$^{4}$ \quad Department of Emergency Medicine, KK Women’s and Children’s Hospital, Singapore 229899, Singapore; e-mail@e-mail.com\\
$^{5}$ \quad
Information Systems, Technology, and Design, Singapore University of Technology and Design, Singapore 487372, Singapore; e-mail@e-mail.com\\}

\abstract{\textcolor{black}{Intelligent systems are transforming the world, as well as our healthcare system. We propose a deep learning-based cough sound classification model that can distinguish between children with healthy versus pathological coughs such as asthma, upper respiratory tract infection (URTI), and lower respiratory tract infection (LRTI). In order to train a deep neural network model, we collected a new dataset of cough sounds, labelled with clinician's diagnosis. The chosen model is a bidirectional long-short term memory network (BiLSTM) based on Mel Frequency Cepstral Coefficients (MFCCs) features. The resulting trained model when trained for classifying two classes of coughs -- healthy or pathology (in general or belonging to a specific respiratory pathology), reaches accuracy exceeding 84\% when classifying cough to the label provided by the physicians' diagnosis. In order to classify subject's respiratory pathology condition, results of multiple cough epochs per subject were combined. The resulting prediction accuracy exceeds 91\% for all three respiratory pathologies. However, when the model is trained to classify and discriminate among the four classes of coughs, overall accuracy dropped: one class of pathological coughs are often misclassified as other. However, if one consider the healthy cough classified as healthy and pathological cough classified to have some kind of pathologies, then the overall accuracy of four class model is above 84\%. A longitudinal study of MFCC feature space when comparing pathologicial and recovered coughs collected from the same subjects revealed the fact that pathological cough irrespective of the underlying conditions occupy the same feature space making it harder to differentiate only using MFCC features.}}


\begin{document}



\section{Introduction}

Cough is a prevalent clinical presentation in many childhood respiratory pathologies including asthma, upper and lower respiratory tract infection (URTI and LRTI), atopy, rhinosinusitis and post-infectious cough \cite{shields2008recommendations,shields2013difficult,oren2015cough}. Because of its wide range of aetiologies, the cause of cough can be misdiagnosed and inappropriately treated \cite{shields2008recommendations}. Clinical differentiation for pathological respiratory conditions take into consideration the history of the presenting respiratory symptoms as well as clinical signs such as pyrexia (i.e., raised body temperature), respiratory rate, shortness of breath and chest auscultation of pathognomonic breath sounds. In some cases, additional investigations such as chest radiographs, laboratory blood tests, bronchoscopy and spirometry are required to reach a definitive diagnosis. These investigations often require hospital visits and place demands on healthcare resources. Moreover, such visits may create a negative social economic impact on the ill child and on his/her family (such as time away from work and childcare arrangements). Further, some of these investigations such as chest radiographs, and  blood tests can result in more harm than benefit, if performed indiscriminately.

There is a growing interest in characterizing acoustic features to allow objective classification of cough sounds originating from different respiratory conditions. Previous studies have looked at medical screenings based on cough sounds \cite{abaza2009classification,murata1998discrimination,abeyratne2013cough,swarnkar2019stratifying,schroder2016classification}. \citet{abaza2009classification} analysed the characteristics of airflow and the sound of a healthy cough to train a classifier that distinguishes between healthy subjects and those with some kind of lung disease. Their model incorporates a reconstruction algorithm that uses principal component analysis. It obtained an accuracy of 94\% and 97\%  to identify abnormal lung physiology in female and male subjects, respectively. \citet{murata1998discrimination} used time expanded wave forms combined with spectrograms to differentiate between productive (i.e., coughs producing phlegm) and non-productive coughs (i.e., dry coughs). Cough sound analysis has also been used to diagnose pneumonia \cite{abeyratne2013cough} and \citet{swarnkar2019stratifying} used it to assess the severity of acute asthma. The latter reported that their model can predict  between children suffering from breathing difficulties involving acute asthma and can characterize the severity of airway constriction. In \cite{botha2018detection}, tuberculosis (TB) screening was investigated using short-term spectral information extracted from cough sounds. They reported an accuracy of 78\% when distinguishing between coughs of TB positive patients and healthy control group. Further, it was noted that the TB screening accuracy increased to 82\% when clinical measurements were included along with features extracted from cough audio. The cough sounds used in some of the aforementioned investigations were carefully recorded in studio environments, whereas the database used in this investigation is collected using a smartphone in a real hospital setting (see Section 2). This type of ecological data collection (or unconstrained audio collection) is of more practical use to physicians, and may also help in developing a mobile phone app in the future that will be more robust when performing early diagnosis of respiratory tract infections in a real-life setting.

There are some studies that use a realistic cough sound database: A Gabor filterbank (GFB)  \cite{schroder2016classification} was used to classify coughs sounds as being `dry' or `productive'.  They reported an accuracy of more than 80\% when incorporating acoustic cough data collected through a public telephone hotline. Another study reported a similar accuracy in classifying wet and dry cough sounds,  though the data was collected using a smartphone \cite{nemati2019comprehensive}.  Recently, this strategy of collecting cough sounds is getting popular \cite{sharma2020coswara,cohen2020novel,orlandic2020coughvid}. Such audio- based strategy  have profound implication when examining symptomatic cough sounds associated with  COVID-19 whereby cough is a primary symptom, alongside fever and fatigue. Convolution Neural Network (CNN) based systems were trained to detect cough and screen for COVID-19, and reported accuracy exceeding 90\% in \cite{wei2020real,imran2020ai4covid,laguarta2020covid} and  while another study had reported 75\% accuracy \cite{bagad2020cough}.  Features were extracted (both handcrafted and transfer learned) from a crowd-sourced database containing breathing and cough sounds \cite{brown2020exploring} and were used to train a support vector machine and ensemble classifiers to screen COVID-19 individuals from healthy controls. They reported an accuracy around 80\%.

There is another line of research inquiry which mainly focus on cough event detection (i.e., to  identify the presence of cough events) in audio recordings \cite{wang2015audio, barry2006automatic, stegmaier1995cough, amoh2015deepcough,nemati2018private,tracey2011cough}, however, in this investigation, we manually segment the cough epochs, and thus review of such studies are outside scope of this report. The aim of this study is to determine if a predictive machine learning model, trained using acoustic features extracted from cough sounds, could be a useful classifier to differentiate common pathological cough sounds from healthy-voluntary coughs (i.e., cough sounds collected from healthy volunteers). The knowledge gained through such methods, could support with the early recognition and triage of medical care, as well as assisting physicians in the clinical management which includes making a differential screening and monitoring of the health status in response to medical interventions.

In the authors' earlier work, audio-based cough classification using machine learning has shown to be a potentially useful technique to assist in differentiating asthmatic cough sounds from healthy-voluntary cough sounds in children \cite{hee2019development,bt2020asthmatic}. The current paper builds upon this previous work (the earlier one used a simple Gaussian Mixture Model- Universal Background Model (GMM-UBM) \cite{hee2019development,bt2020asthmatic}) and uses the collected cough sound dataset to train a deep neural network (DNN) model that can differentiate between pathological and healthy-voluntary subjects. The proposed deep neural network model is trained using acoustic features extracted from the cough sounds. Three different pathological conditions were considered in this investigation: asthma, upper respiratory tract infection (URTI) and lower respiratory tract infection (LRTI). The accuracy of the proposed trained model is evaluated by comparing their predictions against the clinician's diagnosis.

\section{Data Collection}

\subsection{Subject Recruitment}

Subjects in this study was divided into 2 cohorts: Healthy cohort (without respiratory conditions) and the pathological cohort  (with respiratory conditions which included LRTI, URTI and Asthma. LRTI included a spectrum of respiratory diseases labelled as bronchiolitis, bronchitis, bronchopneumonia, pneumonia, lower respiratory tract infection). Participants were recruited from KK Children’s Hospital, Singapore. Inclusion criteria in the pathological cohort was the presence of concomitant symptom of cough, while  inclusion criteria for the healthy cohort was absence of active cough and active respiratory conditions. Pathological cohorts were recruited from the Children’s Emergency Department, Respiratory Ward, and Respiratory Clinic. The cough sounds were recorded during their initial presentation at the hospital. The healthy cohorts were recruited from the Children Surgical Unit. These healthy children were first screened by the anaesthetic team and recruited for the study.


\subsection{Cough Dataset}
\label{Section:Coughdata}
A smartphone was used to record cough sounds from both pathological and healthy children (i.e., without respiratory conditions). For both groups, the subjects were instructed to cough actively. This often resulted in multiple cough epochs per participant (on average 10 to 12).  Recordings were collected at a sampling rate of 44.1 kHz in an unconstrained setting, i.e., a hospital ambience with background noise such as  talking in background, beeping sounds from monitoring devices, alarm sounds, ambulance siren, etc. The collected cough audio files were manually segmented into individual coughs so as to form different entries in the dataset. Characteristics of the resulting dataset are shown in Table~\ref{tab:coughdataset_first}. The working diagnosis for the aetiology of the cough was determined by the clinician based on the clinical history, physical examination, and for some cases investigations such as laboratory tests and chest x-rays were also used.


\begin{table}[H]
\centering
\caption{Characteristics of the collected cough dataset.}
\label{tab:coughdataset_first}
\begin{tabular}{ccccc}
\toprule
\vspace{.1cm}
 & Healthy  & Asthma  & LRTI  & URTI  \\
  &  cohort &  cohort &  cohort &  cohort \\
 \toprule
 Number of Subjects & 89 & 89 & 160 & 78 \\
 \midrule
Number of Coughs & 1149 & 1192 & 2344 & 1240 \\
 \midrule
 Age in Years (SD)&9.07 (2.88)&	8.51 (3.02)&	6.77 (2.65)&	7.21 (2.96)\\
 \midrule
 Gender - &	80 : 9&	60 : 29&	94 : 66&	35 : 43\\
 (Male : Female)&	&&	&	\\
  \midrule
Duration of history &	&&	&	\\
of cough at &NA$^*$	& 3.87 (4.23)&6.63 (5.93)	&5.22 (2.72)	\\
presentation; day (SD) &	&&	&	\\
\bottomrule
\end{tabular}

\end{table}
NA$^*$ - Not Applicable








\section{Trained Models}
Using the dataset described above, five different classification models based on deep neural networks were built.
\subsection{ \textit{Healthy vs Pathology (2-class) Model}}
The first model (\textit{Healthy vs Pathology (2 class) Model}) was trained to classify whether each cough segmented is a healthy-voluntary cough or pathological. Here, we consider all pathological coughs as one class, named as `pathological cough'.

\subsection{\textit{Healthy vs LRTI Model} , \textit{Healthy vs URTI Model}, \textit{Healthy vs Asthma Model}}
The second set of models (three in total) were trained to classify between healthy-voluntary coughs and a particular respiratory pathology (i.e. one respiratory pathology at a time). \textit{Healthy vs LRTI Model} - was trained to predict whether a cough is healthy-voluntary or from a subject diagnosed with LRTI; \textit{Healthy vs URTI Model} - was trained to predict whether a cough is healthy-voluntary or from a subject diagnosed with URTI; and finally \textit{Healthy vs Asthma Model} - was trained to predict whether a cough is healthy or from a subject diagnosed with Asthma.

\subsection{\textit{Healthy vs Pathology (4-class) Model}}

The final classification model was trained to predict between all the four chosen classes. Thus, \textit{Healthy vs Pathology  (4-class) Model} - classifies whether a cough is healthy-voluntary or associated with any of the three pathological conditions of LRTI, URTI, or asthma.

\section{Classification Model}
\subsection{Long Short Term Memory (LSTM)}
An LSTM based network was chosen as the classification model in this investigation. LSTM networks take sequence-data as the input, and makes predictions based on their sequence dynamic characteristics by learning long-term dependencies between time steps of sequence data. They are known to work well for their ability to handle sequence data due to their memory mechanism \cite{hochreiter1997long}. Our choice for LSTM is motivated by the sequential nature of audio data and their ability to handle input audio features that vary in length \cite{hochreiter1997long,glorot2010understanding}, as is the case with the features extracted from the collected cough sounds (see Section \ref{section:soundprocessing}).

\begin{figure}[ht]
 \centering
 \includegraphics[width=1.0\linewidth]{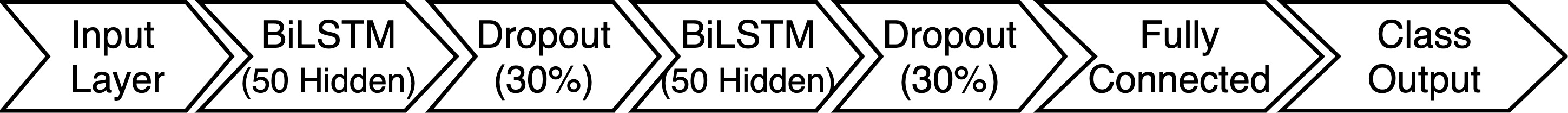}
 \caption{Deep Neural Network Architecture using BiLSTM layers.}
 \label{fig:methodologyarchi}
\end{figure}

In this investigation, we used a four-layer neural network with two deep layers of bidirectional LSTMs (BiLSTMs) (See Figure \ref{fig:methodologyarchi}). Each BiLSTM layer learns bidirectional long-term dependencies from sequence data. These dependencies will help the network to understand the long-term dynamics present in the features and thus learning the complete time series \cite{schuster1997bidirectional,graves2005framewise}. We have investigated different deep neural network types such as fully connected deep neural networks, LSTMs, BiLSTMS, so as to identify the best classification model for our cough screening problem. In the end, BiLSTMS were chosen, as they were found to produce better results for the chosen feature sets (A similar turn around was reported in \cite{graves2005framewise}). These network comparison results are not shown as they are outside the scope of this paper.

\subsection{BiLSTM Architecture}

The first layer (input layer) has a dimension of 42 so as to match the size of the MFCC feature vectors corresponding to every audio frame (see Section \ref{section:soundprocessing}). The second layer is a BiLSTM layer with 50 hidden units. This is followed by a dropout layer which in turn is followed by another BiLSTM and a dropout layer. The second BiLSTM layer also has 50 hidden units. A 30\% dropout was chosen for both of the dropout layers. Finally, depending on the classification objective, we used either two fully connected layers (for the 2-class classification problem) or four fully connected layers (for the 4-class classification problem). The networks were optimized to minimize cross-entropy loss with sigmoid activation. This particular architecture was selected after multiple hyper-parameter optimization steps. We used grid search to find the optimal number of hidden units, the number of hidden layers, as well as the dropout rate. The resulting combination reported in this paper was able to reach the lowest training loss (or in other words maximum training accuracy)  when trained for multiple cough classification hypotheses.


\section{Experimental Methodology}

\subsection{Dataset Split}

The collected dataset was randomly split (70-30\%) into two non-overlapping parts: training and test set. The resulting split sizes are shown in Table \ref{tab:coughsound_data}. We made sure that cough sounds belonging to the same person were either in the test or in the training set, but not in both. Since the test data has not yet been seen by the model during the training phase, one could expect that the resulting performance of this model offers a good approximation for what can be expected in a real scenario (i.e., when the model is asked to make a prediction for an unseen cough).

\begin{table}[h]

\centering
\caption{Number of instances of the cough sounds in the training and test set.}
\begin{tabular}{l cc cc}
\toprule
\multirow{2}{*}{Class} & \multicolumn{2}{c}{Number of Children} & \multicolumn{2}{c}{Number of Coughs} \\

 & Training & Test & Training & Test \\
 \midrule
URTI & 54 & 24 & 849 & 391 \\
LRTI & 113 & 47 & 1,679& 665 \\
Asthma & 65 & 24 & 726 & 466 \\
Healthy & 51 & 38 & 645 & 504 \\
\bottomrule
\end{tabular}
\label{tab:coughsound_data}
\end{table}

\subsection{Methodology}
\begin{figure}[ht]
 \includegraphics[width=1.0\linewidth]{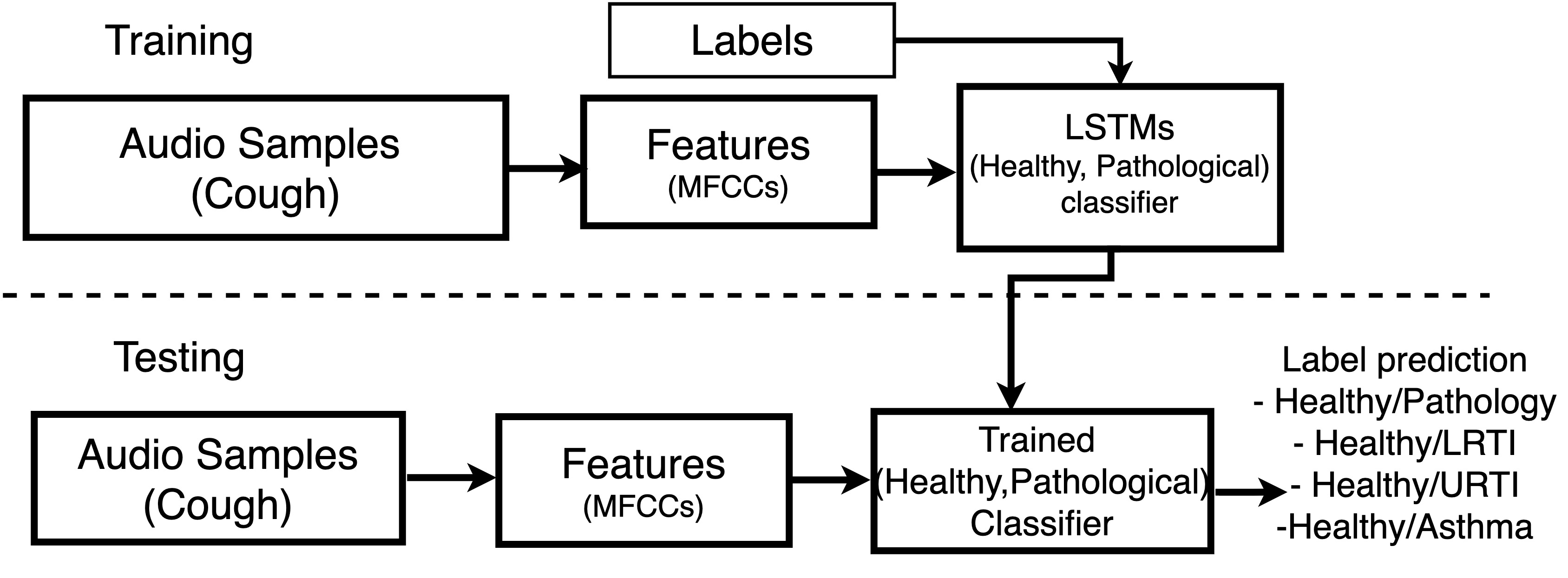}
 \caption{Experimental Methodology followed in our investigation.}
 \label{fig:methodology}
\end{figure}

The general experimental methodology followed in this investigation is shown in Figure~\ref{fig:methodology}. We first trained our deep neural network models using features extracted from data from our training set, and then proceeded to evaluate the models using a separate test set. The trained model is used to predict which class a cough sound belongs to. This cough prediction was subsequently used to screen whether a subject is healthy or having some respiratory conditions. This screening is done based on the most frequent (mode) prediction outcome of all the cough sounds belonging to a particular subject. In what follows, we discuss how the data has been pre-processed, which audio features were chosen for this investigation, and how the model was built.

\subsection{Cough Sound Processing \& Audio Feature Extraction}
\label{section:soundprocessing}
The segmented cough sounds were detrended to remove any linear trends, baseline shifts, or slow drifts, then normalized (to have a maximum sample value of one), and finally downsampled (downsampled to 11.025 kHz from the original sampling rate of 44.1 kHz).


The pre-processed audio signals were first segmented into frames of 100 ms, after which a Hamming window was applied, followed by the extraction of audio features. Mel-frequency Cepstral Coefficients (MFCCs) were chosen for this investigation owing to their effectiveness when it comes to audio classification problems \cite{balu2019spoofing,muda2010voice}. MFCCs are a set of features that focus on the perceptually relevant aspects of the audio spectrum, additionally the coefficients could contain information about the vocal tract characteristics \cite{rabiner2011theory,kawakami2014speaker}. In this investigation we used 14 MFCCs with their deltas and delta-deltas, thus resulting in a total of 42 coefficients (14 MFCCs, 14 deltas and 14 delta-deltas) for every audio frame.


\subsection{Measuring performance}
The performance of  DNN models is measured by calculating the classification accuracy and is further analyzed using the receiver operating characteristic (ROC) \cite{brown2006receiver}. and confusion matrix \cite{tarca2007machine}.

\subsubsection{Accuracy}
The classification accuracy is calculated by comparing the predicted outputs with the actual outputs.
\begin{equation}
\textit{Accuracy}=\frac{\textit{Number of correct predictions}}{\textit{Total number of predictions}}
\end{equation}

\subsubsection{Receiver operating characteristic (ROC)}
The ROC is created by plotting the true positive rates (i.e., sensitivity: the ratio of true positives over the sum of true positives and false negatives) against the false positive rates (i.e., 100 \textit{minus} specificity; specificity is the ratio of true negatives over the sum of false positives and true negatives) for various decision thresholds. A perfect model results in a ROC curve which passes close to the upper left corner, indicating a higher overall accuracy. This would thus result in a ROC of which the area underneath (AROC) equals 1.

\subsubsection{Confusion matrix}
The performance of a classifier was further analysed using confusion matrices, whereby the true and false positives and negatives are displayed for each class. For a good classifier, the resulting confusion matrix will have large numbers along the diagonal (i.e., values closer to 100\%). The percentage of misclassified data is reflected in the off-diagonal elements.

\section{Results}

\subsection{Power Spectrum Comparison}

From the original cough sounds, the power spectrum (i.e., the distribution of energy contained within the signal over various frequencies) was estimated. These frequencies were then grouped into five equal bins between 0 to $f_s/2$ (whereby $f_s$ is the sampling frequency) and the corresponding spectral power present in each of these bins was calculated.

 \begin{figure}[h!]
 \centering
 \includegraphics[width=1.0\linewidth]{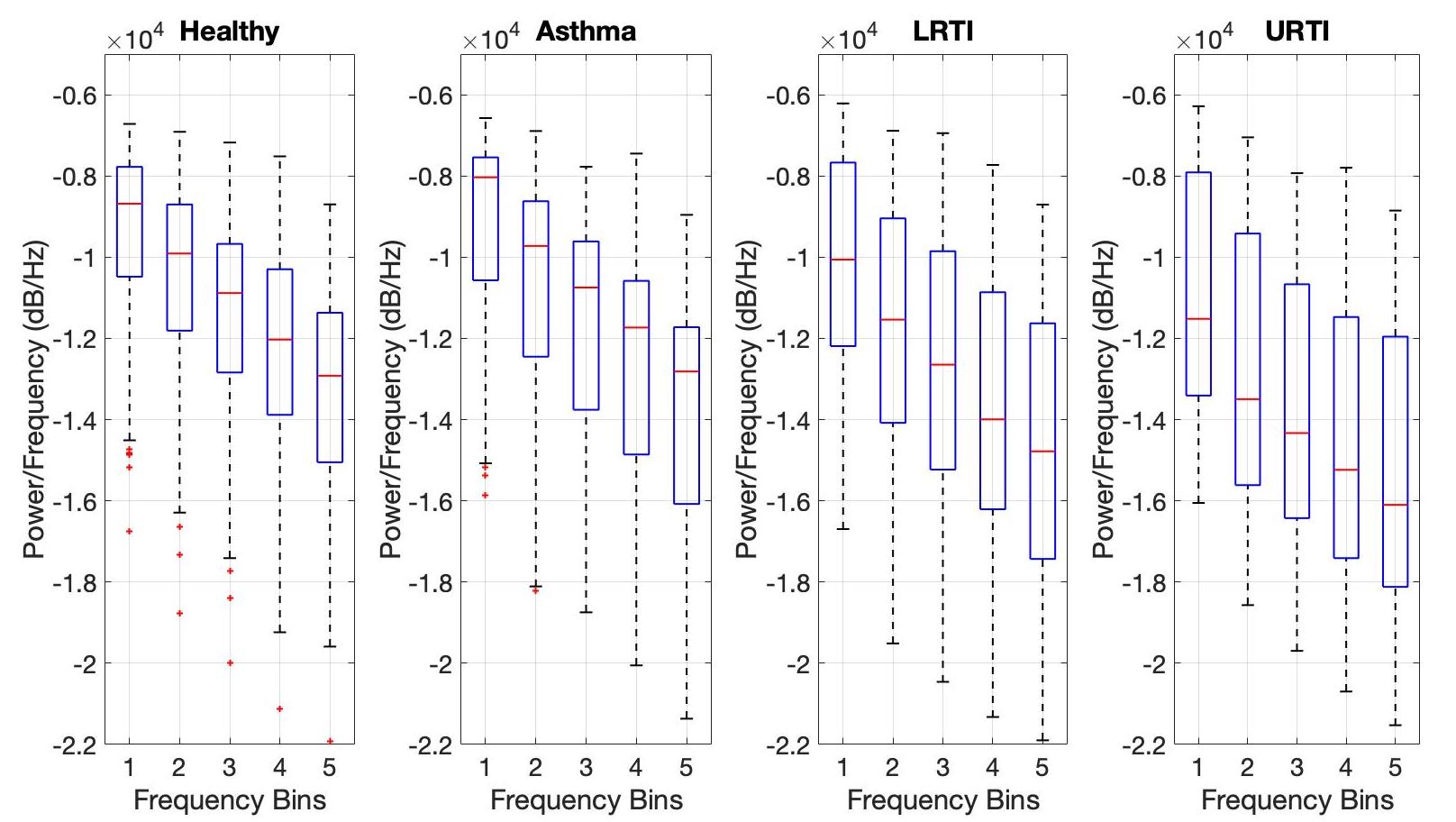}
 \caption{Boxplot showing the distribution of the power spectrum across various frequency bins.  Bin 1 -- 0 to 1.1 kHz, Bin 2 -- 1.1 to 2.2 kHz, Bin 3 -- 2.2 to 3.3 kHz, Bin 4 -- 3.3 to 4.4 kHz, Bin 5 -- 4.4 to 5.5 kHz}
 \label{fig:Spectral}
\end{figure}

 The distribution of the power spectrum for 500 randomly chosen cough samples (of different respiratory conditions) is shown using a boxplot (See Figure \ref{fig:Spectral}). The median is shown using a red line. The bottom and top edges of each of the boxes indicate the $25^{th}$ and $75^{th}$ percentile, respectively. The likely range of variation (i.e., inter-quartile range (IQR)) is given by distances between the tops and bottoms \cite{mcgill1978variations}.

 The median line corresponding to every bin (for both the healthy and pathological coughs) does not appear to be centred inside the box (i.e., the possible mean of each bin), thus indicating that the power distribution is slightly skewed for each bin. IQR is found to be slightly larger in spectral power bins of pathological cough when compared to the healthy spectral bin. Overall, there are no clear trends between the median value of the spectral bin for healthy and pathological coughs. The asthmatic spectral bins tend to have a slightly higher median value compared to the spectral bins of healthy coughs. The opposite trend was found when comparing spectral bins of LRTI and URTI against that of healthy. We speculate that this may be due to the fact that both these conditions (LRTI and URTI) include inflamed airway tissues, which may increase acoustic damping (especially at high frequency). This postulate requires further investigation. In addition, the difference observed maybe attributed to variability in subject characteristics between the groups such as age, gender between groups.  See  Table~\ref{tab:coughdataset_first}.


\subsection{Feature Analysis -- MFCCs -- Extracted for investigation}
\label{Featurestudy}
\begin{figure}[h!]
 \centering
 \includegraphics[width=1.0\linewidth]{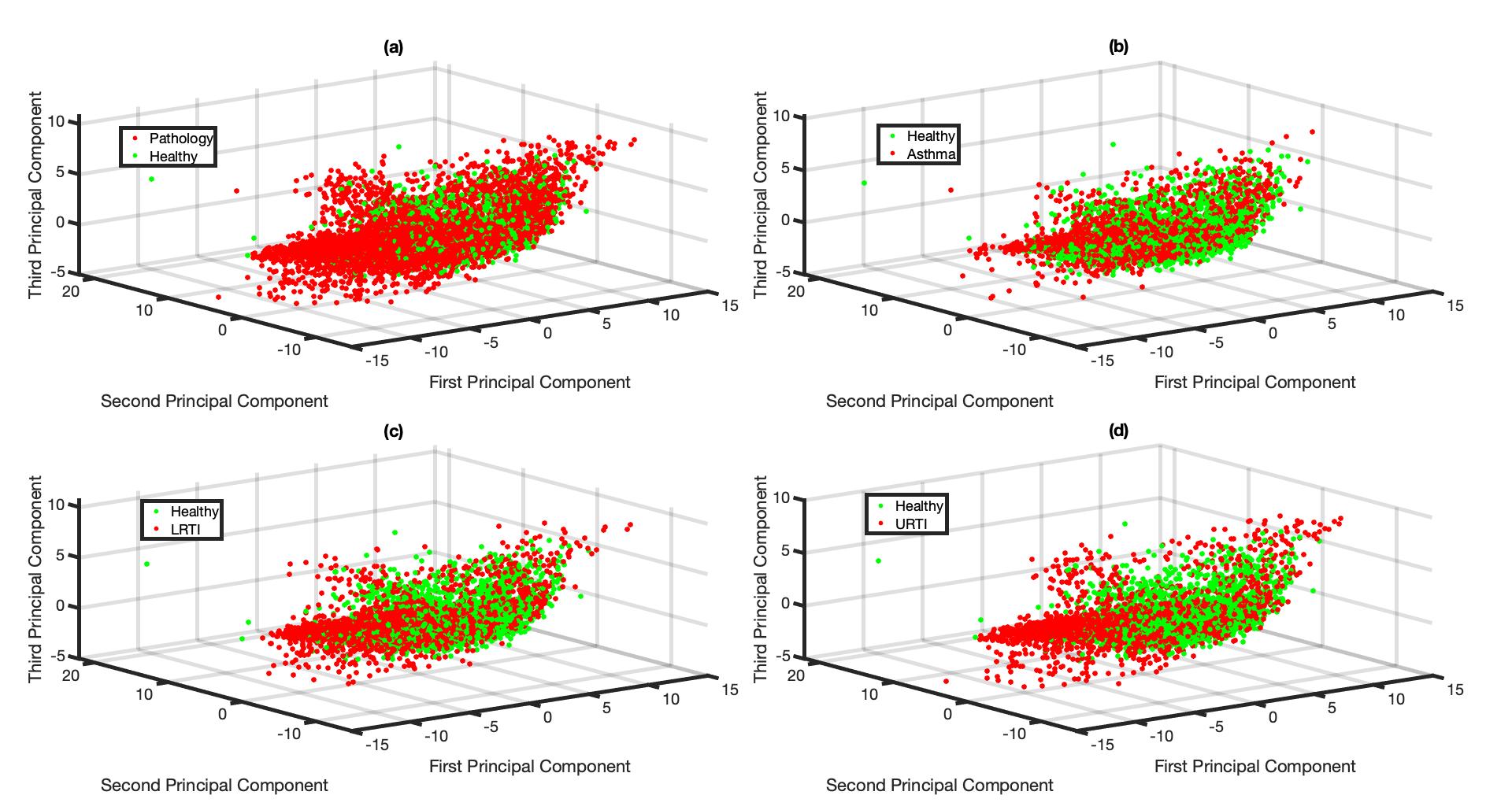}
 \caption{MFCC feature visualization after transforming the original 42 dimensions to 3 dimensions using PCA. (a) Healthy vs Pathology. (b) Healthy vs LRTI. (c) Healthy vs URTI (d) Healthy vs Asthma. }
 \label{fig:PCA}
\end{figure}

The objective of this feature analysis is to understand if cough sounds contain any subtle cues to distinguish between healthy and pathological subjects. The higher dimensional MFCC features extracted from various respiratory pathological coughs were compared against the healthy coughs after transforming them to a lower dimension using Principal Component Analysis (PCA) \cite{wold1987principal}. Such dimensionality reduction techniques often give some insight into the feature space of the chosen classes. The resulting visualisation of the first three PCA components (the first three principal components corresponds to the largest three eigen values and capture more than 95\% of the variance (information) in this dataset) is shown in Figure \ref{fig:PCA}. MFCCs extracted from 5,000 audio frames from each of the categories were used for this visualization. All of these audio frames were part of the training set used for training the BiLSTM network.

No clear clusters are visible in the feature space (See Figure \ref{fig:PCA}). This is true for all the four investigated cases: features of healthy versus pathological cough sound signal, and features of healthy coughs when compared to features from each individual respiratory pathologies (See Figure \ref{fig:PCA}(a), (b), (c) and (d)). This reflects the anecdotal observations that clinicians themselves find it hard to distinguish  these pathologies based on cough sound alone.

\subsection{Feature Analysis -- MFCCs -- Longitudinal Study}
\label{longitudestudy}
 The objective of this longitudinal study is to understand the evolution of the feature space of MFCCs over time for the different classes of respiratory conditions. For this study, the cough sounds were collected and organised in a two-stage process. In the first stage, 51 subjects recruited from the hospital were asked to make multiple voluntary cough sounds (on average 10 to 12 coughs). There were 24 subjects with Asthma, seven with URTI and 20 with LRTI. In the second stage, these 51 subjects were were followed up upon recovery after hospital discharge (approximately two weeks after hospital discharge) and voluntary cough sounds (on average 10 to 12) were again collected. It is important to note here that Stage 1 coughs were a part of the cough dataset used for training the BiLSTM model, however, Stage 2 coughs were not used in any training process. The cough sounds were recorded as described in Section \ref{Section:Coughdata}.

 \begin{figure}[h!]
 \centering
 \includegraphics[width=1.0\linewidth]{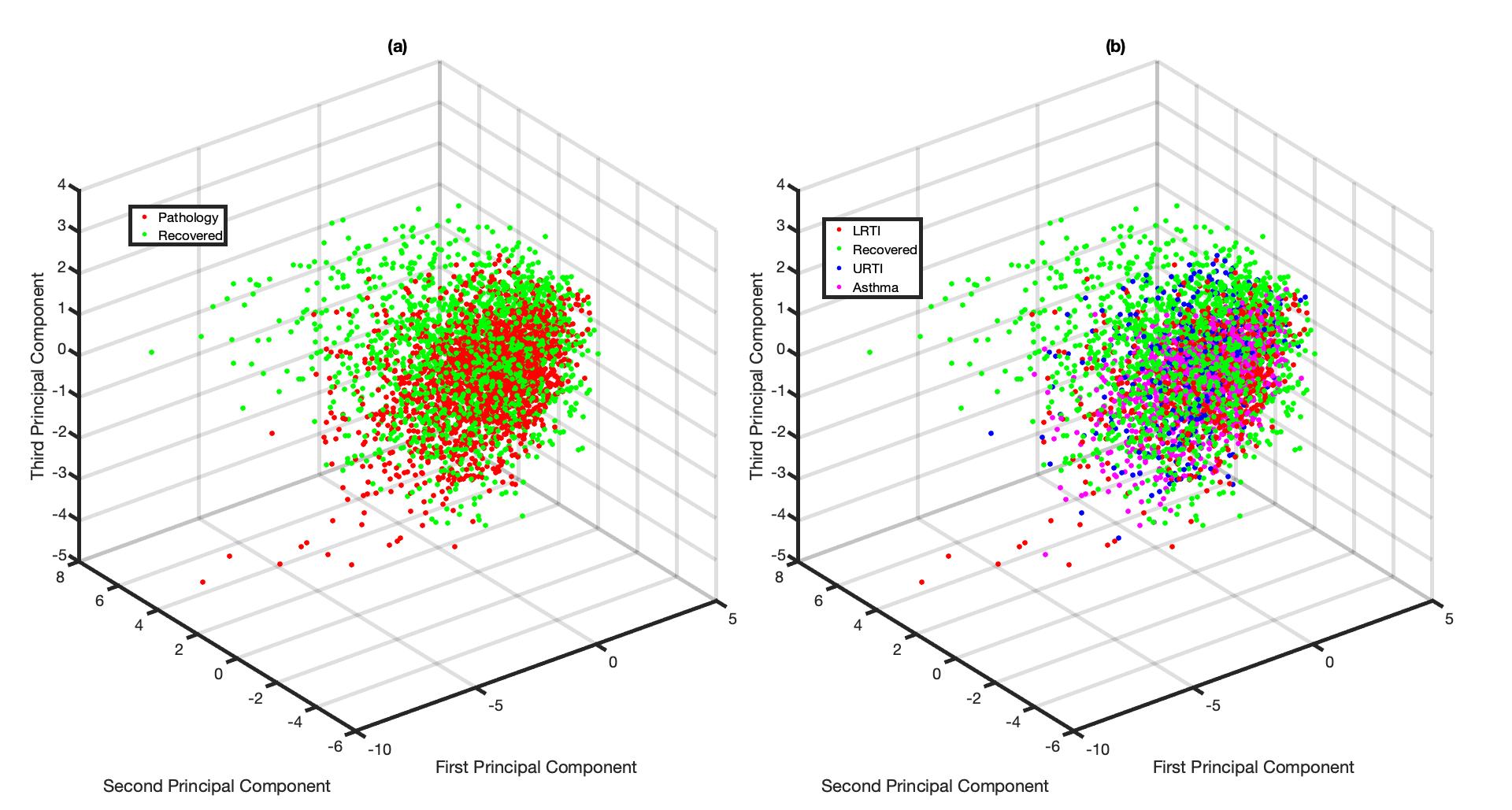}
 \caption{MFCC feature evolution from pathological to recovery. The plot is created after transforming the original 42 dimensions into three using PCA. (a) Recovered vs Pathology. (b) Recovered vs LRTI vs URTI vs Asthma. }
 \label{fig:PCA_longi}
\end{figure}

MFCC coefficients were extracted from the coughs collected from these 51 subjects as described in Section \ref{section:soundprocessing}. There were a total of 3,810 frames analysed as part of this longitudinal study: 1,675 - recovered, 746 - LRTI, 399 - URTI and 990 - Asthmatic. The extracted MFCC coefficients' dimensionalities were then  reduced using PCA for visualization purposes (See Figure \ref{fig:PCA_longi}). Stage 1 coughs can be considered as pathological whereas Stage 2 coughs (i.e., recovered) can be considered to represent healthy-voluntary coughs. The evolution of the MFCC feature space is explored here, since the coughs were collected from the same subject over a period of time.
Just like in Figure \ref{fig:PCA}, no clear clusters are visible when analysing evolution of the extracted features (see Figure \ref{fig:PCA_longi}). Also, it can be seen that MFCC features extracted from Stage 1 coughs occupy relatively the same feature space irrespective of the underlying respiratory conditions (See Figure \ref{fig:PCA_longi}(b)).

With no clear clusters visible in the feature space analysis discussed in Section \ref{Featurestudy} and \ref{longitudestudy}, our classification problem may require the introduction of non-linearity, so as to uncover more complex, hidden, relationships. And thus another motivation for choosing a deep  neural network.


\subsection{Model Performance}
\label{Sec:Modelperfom}
\subsubsection{ \textit{Healthy vs Pathology Model}}

The cough classification accuracy (i.e., accuracy in classifying each cough segment) and the healthy-pathology classification accuracy (i.e., accuracy in classifying entire cough epochs to a particular respiratory pathology) on our test set are shown in Table \ref{tab:classaccura_model1}. The BiLSTM has resulted in good performance when classifying the pathological cough sounds from healthy-voluntary cough sounds, with an accuracy of 84.5\%. Further, when respiratory pathology classification of subject  were made (by considering the entire cough epochs) based on the most frequent (mode) prediction outcome of coughs from a subject for an entire cough epochs, the accuracy gets even higher (91.2\%). This is to be expected,  e.g., if one assumes there are $n$ coughs available per subject, even though model can misclassify individual cough sounds, the respiratory pathological classification result will be wrong only when $(n/2) + 1$ out of the $n$ coughs belonging to a particular patient are misclassified (or in other words  respiratory pathological classification is more robust). Given an accuracy rate of 84.5\% for individual cough prediction, this would be very rare.


\begin{table}[h]
\small
\centering
\caption{ Accuracy of the \textit{  Healthy vs. Pathology Model.}}
\label{tab:classaccura_model1}
\begin{tabular}{l|cc}
\toprule
Model & \begin{tabular}[c]{@{}c@{}}Individual Cough \\ Classification \\ Accuracy \\(in \%) \end{tabular} & \begin{tabular}[c]{@{}c@{}}Respiratory \\Pathology \\ Classification of subject \\ accuracy \\ based on\\  entire cough epoch\\(in \%) \end{tabular} \\
\midrule
\textit{ Healthy vs pathology Model} & 84.5 & 91.2 \\
\\
\hline

\end{tabular}
\end{table}

\begin{figure}[H]
\centering
 {\includegraphics[width=1\linewidth]{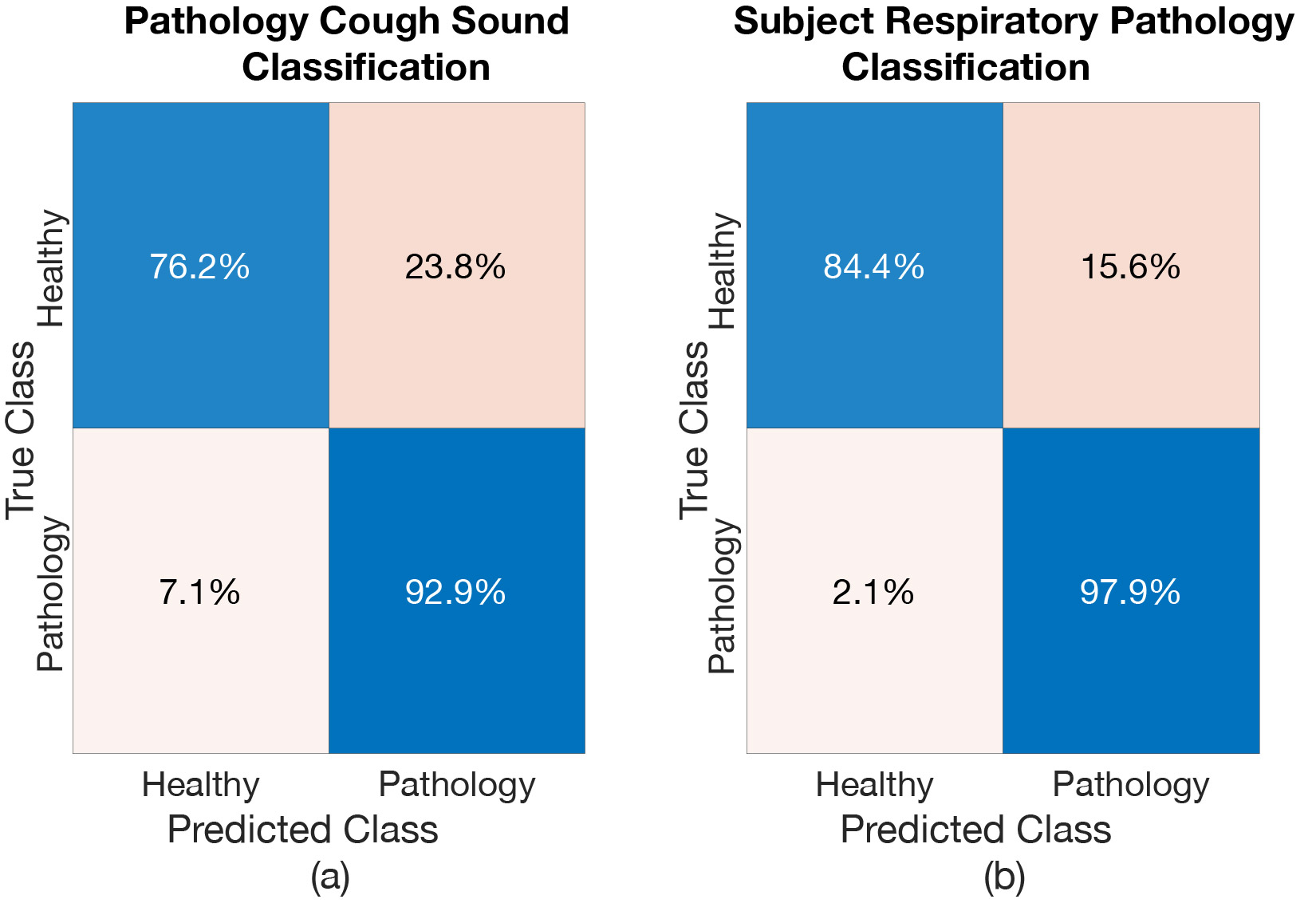}
 }

\caption{Confusion Matrix of \textit{Model Healthy vs pathology} (a) when classifying coughs (b) when classifying subject for respiratory pathology. }
\label{fig:Conf_model1}
\end{figure}

A confusion matrix was created to further analyse the results of this model, see Figure \ref{fig:Conf_model1}. The percentage of healthy-voluntary coughs misclassified as pathological coughs is higher compared to pathological coughs misclassified as healthy-voluntary coughs (23.8\% misclassified compared to 7.1\%, see Figure \ref{fig:Conf_model1}(a)). This higher healthy-voluntary cough misclassification rate further resulted in a relatively large number of healthy subjects misclassified as having a pathology (15.6 \% subjects were misclassified, See Figure \ref{fig:Conf_model1}(b)).

\begin{figure}[H]
\centering
{\includegraphics[width=1\linewidth]{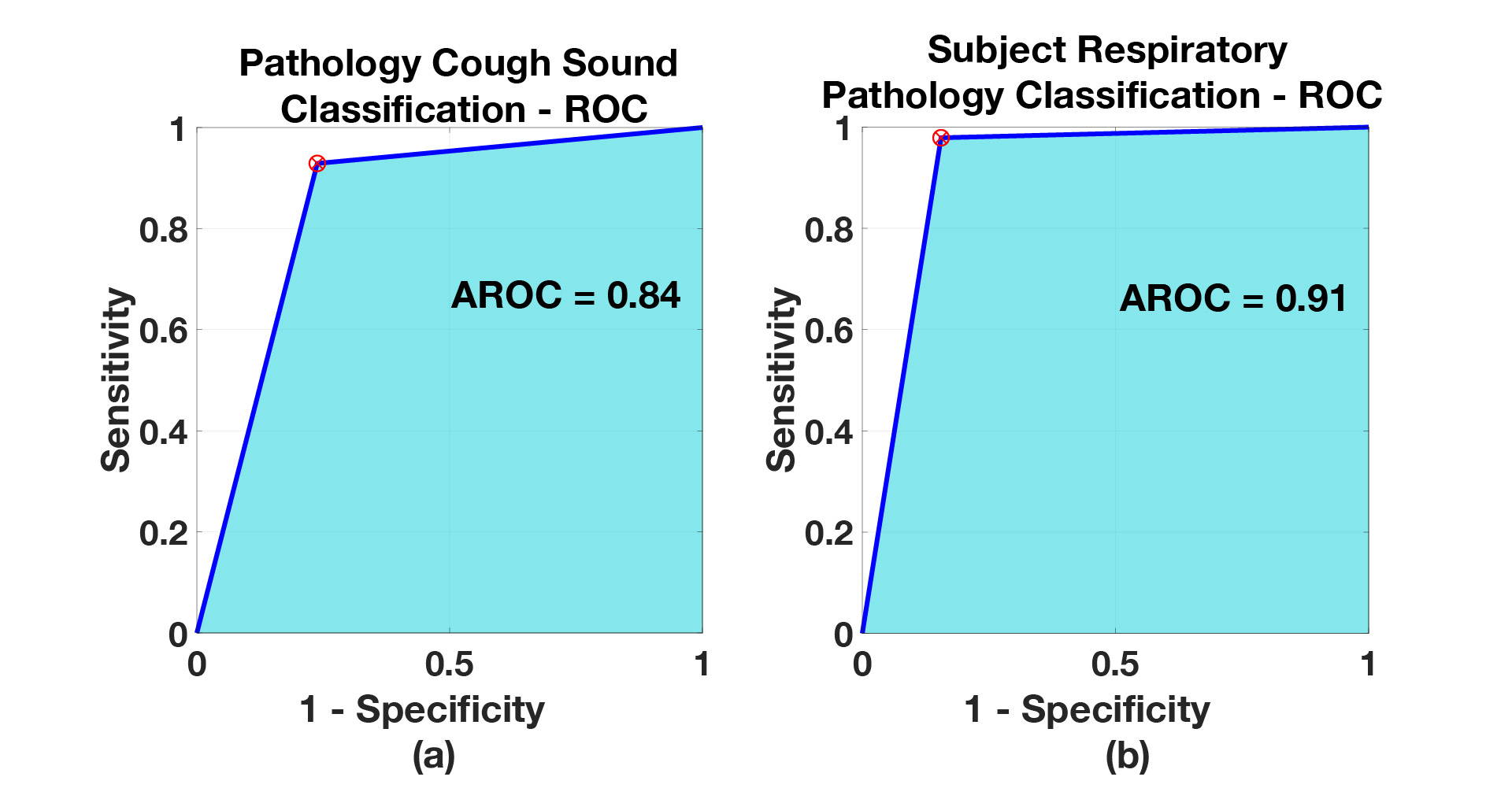}}
\caption{ROC of \textit{Model Healthy vs pathology}. (a) when classifying coughs (b) when classifying subject for respiratory pathology.}
\label{fig:PathresultROC}
\end{figure}

The receiver operating characteristic of this model is shown in Figure \ref{fig:PathresultROC}, along with the corresponding AROC value. The resulting AROC values are 0.84 for cough classification and 0.91 for respiratory pathology classification of subject, see Table \ref{tab:classAROC_model1}). The AROC is convincingly high, which means that the model has delivered  good separability between two classes. Also shown in Figure \ref{fig:PathresultROC}, is the optimum threshold, co-located in the nearest point to (0,1), which maximizes the sensitivity and specificity values (shown as a red cross).

\begin{table}[h]
\small
\centering
\caption{ AROC of \textit{Healthy vs pathology model.}}
\label{tab:classAROC_model1}
\begin{tabular}{l|cc}
\toprule
Model & \begin{tabular}[c]{@{}c@{}}Individual Cough \\ Classification \\ AROC  \end{tabular} & \begin{tabular}[c]{@{}c@{}}Respiratory \\Pathology \\ Classification of subject \\ AROC \\ based on\\  entire cough epoch \end{tabular} \\
\midrule
\textit{Healthy vs pathology Model} & 0.84 & 0.91 \\
\\
\hline

\end{tabular}
\end{table}

\subsubsection{\textit{Healthy vs LRTI Model}, \textit{ Healthy vs URTI Model}, \textit{ Healthy vs Asthma Model}}

The resulting cough classification accuracy and the respiratory pathology classification of subject accuracy when considering one respiratory pathology at a time is shown in Table \ref{tab:classaccura_pathoone}. Again, the deep BiLSTM was able to produce good results when differentiating the healthy-voluntary coughs from those resulting from various respiratory conditions. This resulted in classification accuracy exceeding 85\% for every investigated scenario. Respiratory pathology classification of subjects, as expected, result in even higher accuracy (exceeding 92\% for every case).

\begin{table}[th]
\small
\centering
\caption{Experimental results in terms of accuracy.}
\label{tab:classaccura_pathoone}
\begin{tabular}{l|cc}
\toprule

Model & \begin{tabular}[c]{@{}c@{}}Individual Cough \\ Classification \\ Accuracy \\(in \%) \end{tabular} & \begin{tabular}[c]{@{}c@{}}Respiratory \\Pathology \\ Classification of subject \\ accuracy \\ based on\\  entire cough epoch\\(in \%) \end{tabular} \\
\midrule
\textit{Model Healthy vs LRTI} & 86.3 & 94.5 \\
\textit{Model Healthy vs URTI} & 86.5 & 92.7  \\
\textit{Model Healthy vs Asthma} & 85.9 & 94.2 \\
\bottomrule
\end{tabular}
\end{table}

\begin{figure}
\centering
{\includegraphics[width=1\linewidth]{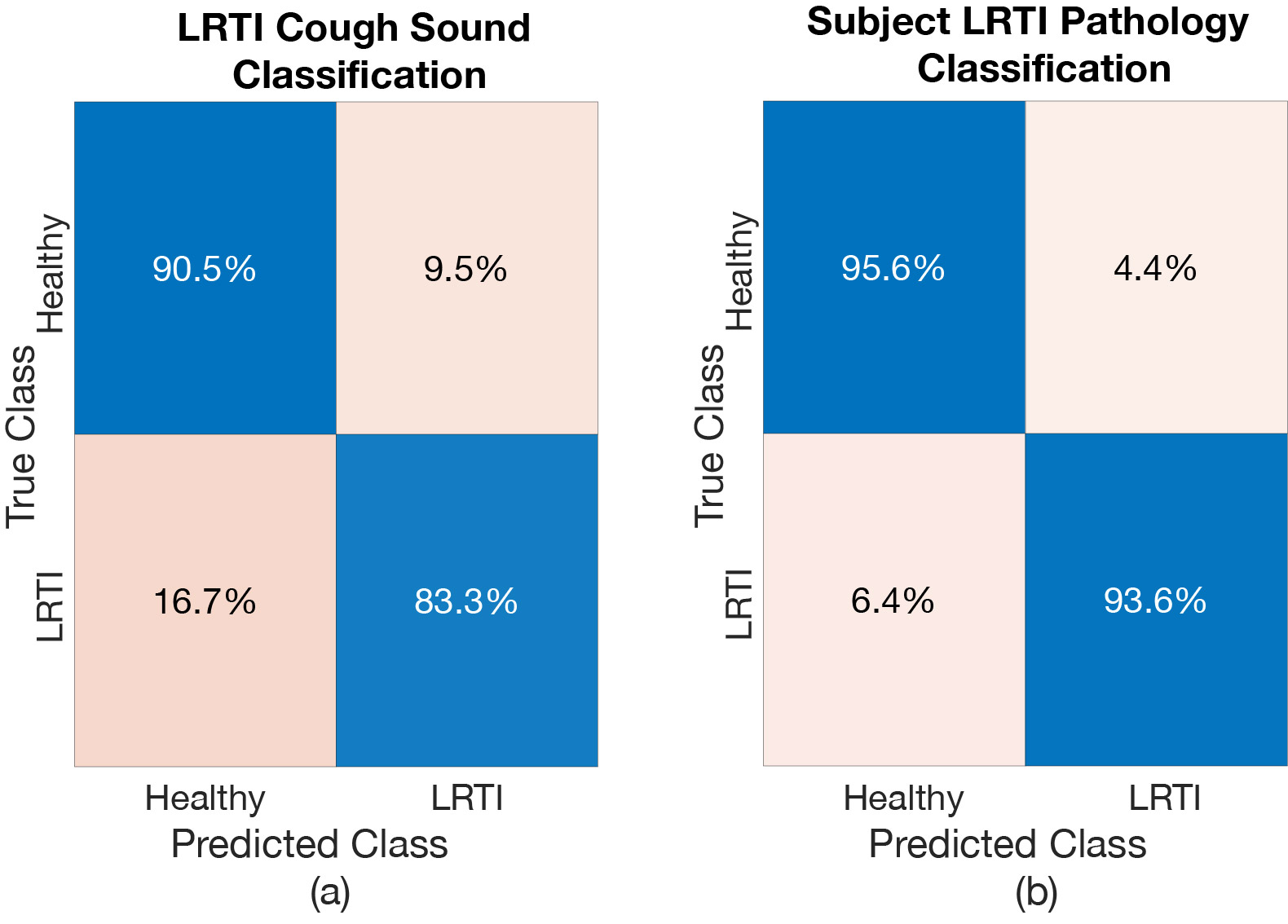}}
\caption{Confusion Matrix of \textit{Healthy vs LRTI Model}.  (a) when classifying LRTI coughs (b) when classifying subject for LRTI. }
\label{fig:LRTIresultconf}
\end{figure}

\begin{figure}[H]
\centering
{\includegraphics[width=1\linewidth]{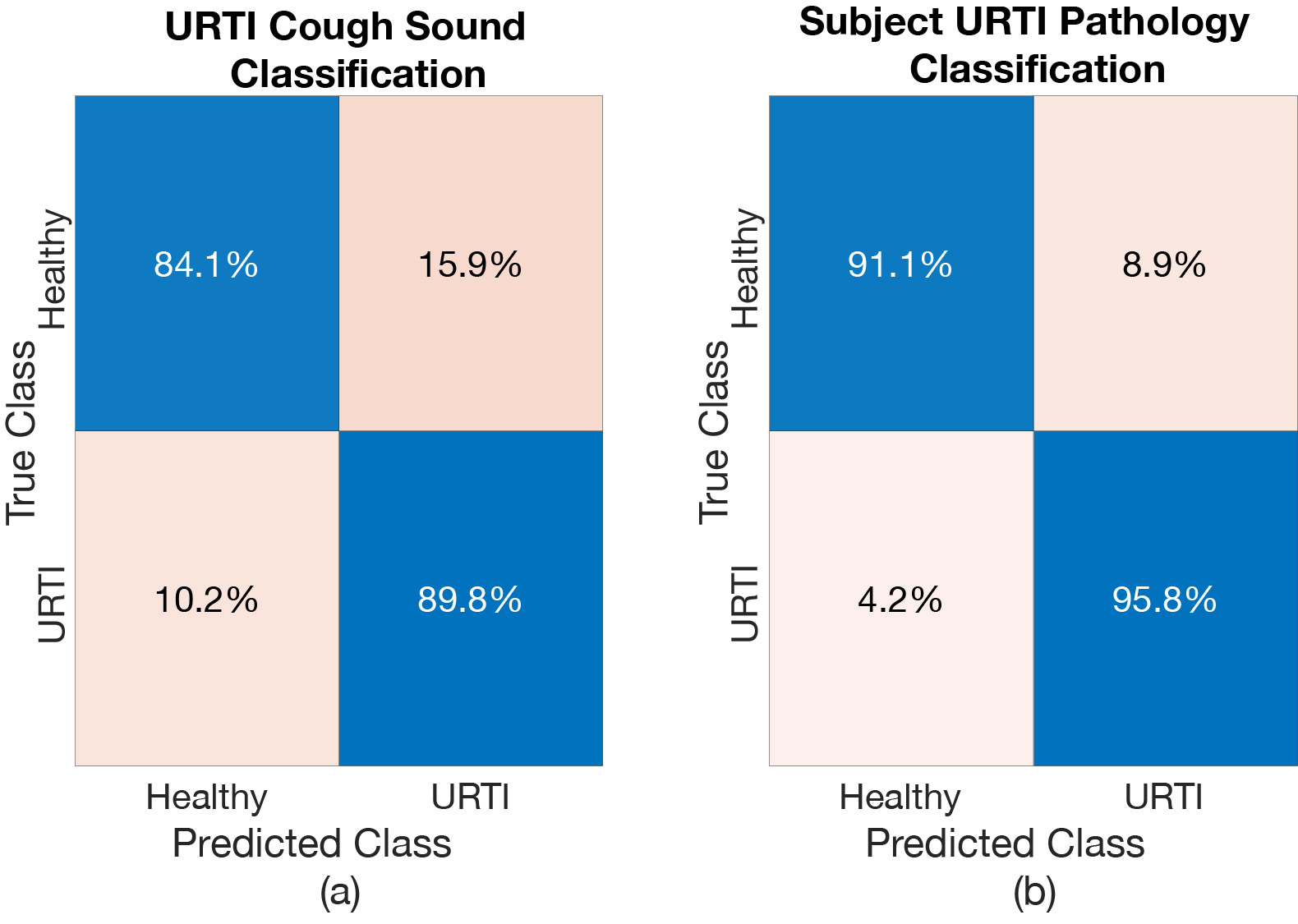}}
\caption{Confusion Matrix of \textit{Healthy vs URTI Model}. (a) when classifying URTI coughs (b) when classifying subject for URTI.}
\label{fig:URTIresultconf}
\end{figure}

\begin{figure}[H]
\centering
 {\includegraphics[width=1\linewidth]{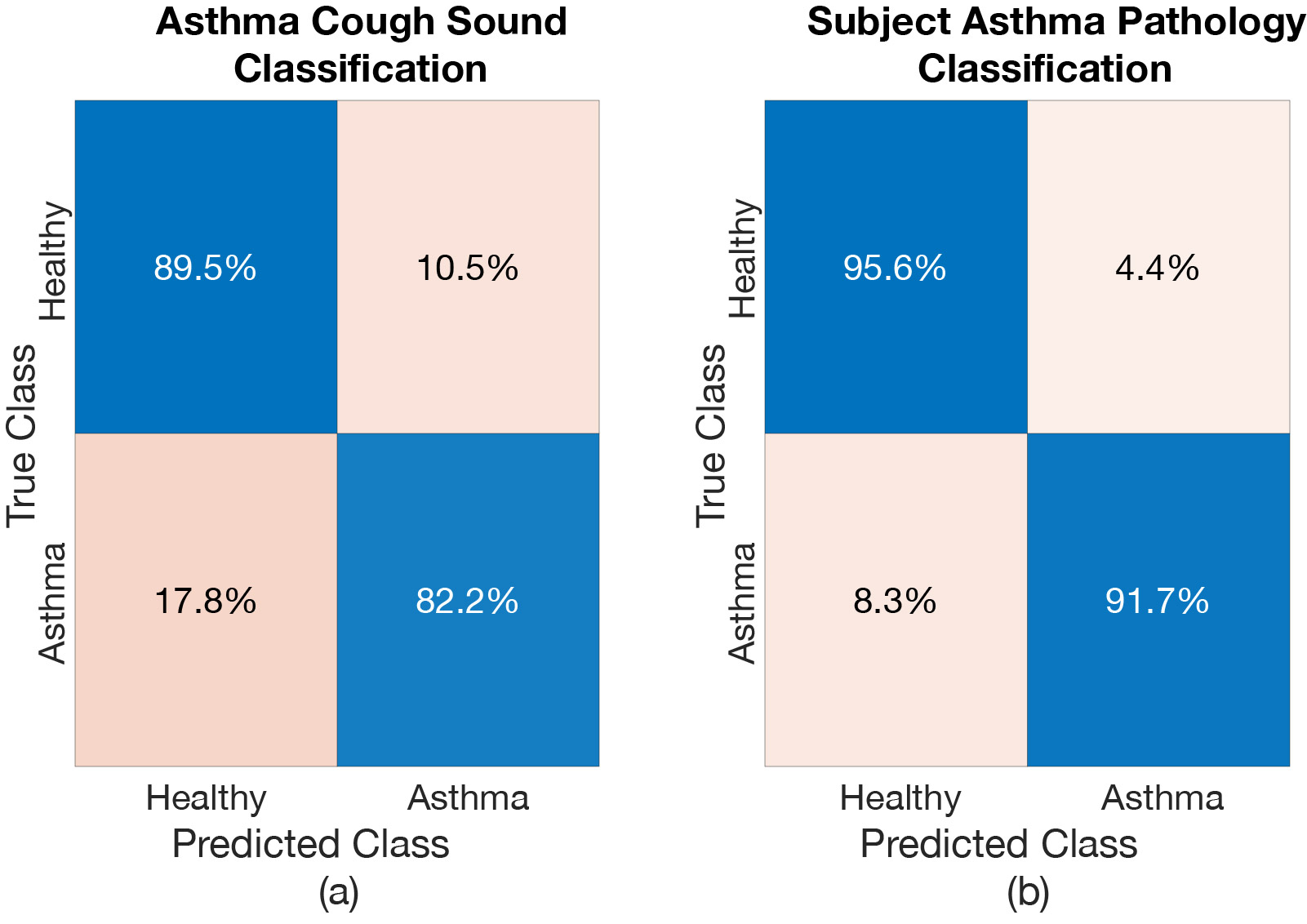}}

\caption{Confusion Matrix of \textit{Healthy vs Asthma Model} (a) when classifying Asthmatic coughs (b) when classifying subject for Asthma. }
\label{fig:Asthmaresultconf}
\end{figure}

Confusion matrices were produced to further analyse the results from each of these models. Figures \ref{fig:LRTIresultconf}, \ref{fig:URTIresultconf} and \ref{fig:Asthmaresultconf} show the confusion matrices for \textit{Healthy vs LRTI Model}, \textit{Healthy vs URTI Model} and \textit{Healthy vs Asthma Model}, respectively. The performance of \textit{Healthy vs LRTI Model} and \textit{Healthy vs Asthma Model} when it comes to correctly classifying healthy coughs from pathological coughs is comparable (see Figure \ref{fig:LRTIresultconf} and \ref{fig:Asthmaresultconf} (a)). \textit{Healthy vs URTI Model} has a slightly larger number of misclassifications when predicting healthy coughs, however, its performance on pathological coughs detection (URTI in this case) is better compared to the other two models (see Figure \ref{fig:URTIresultconf}). When it comes to respiratory pathology classification of subject based on the entire cough epochs, as expected, the classification models have resulted in higher correct classification rate compared to the individual cough classification model (See Figure.,  \ref{fig:LRTIresultconf}, \ref{fig:URTIresultconf} and \ref{fig:Asthmaresultconf} (b)).

\begin{figure}[H]
 \centering
 \includegraphics[width=1\linewidth]{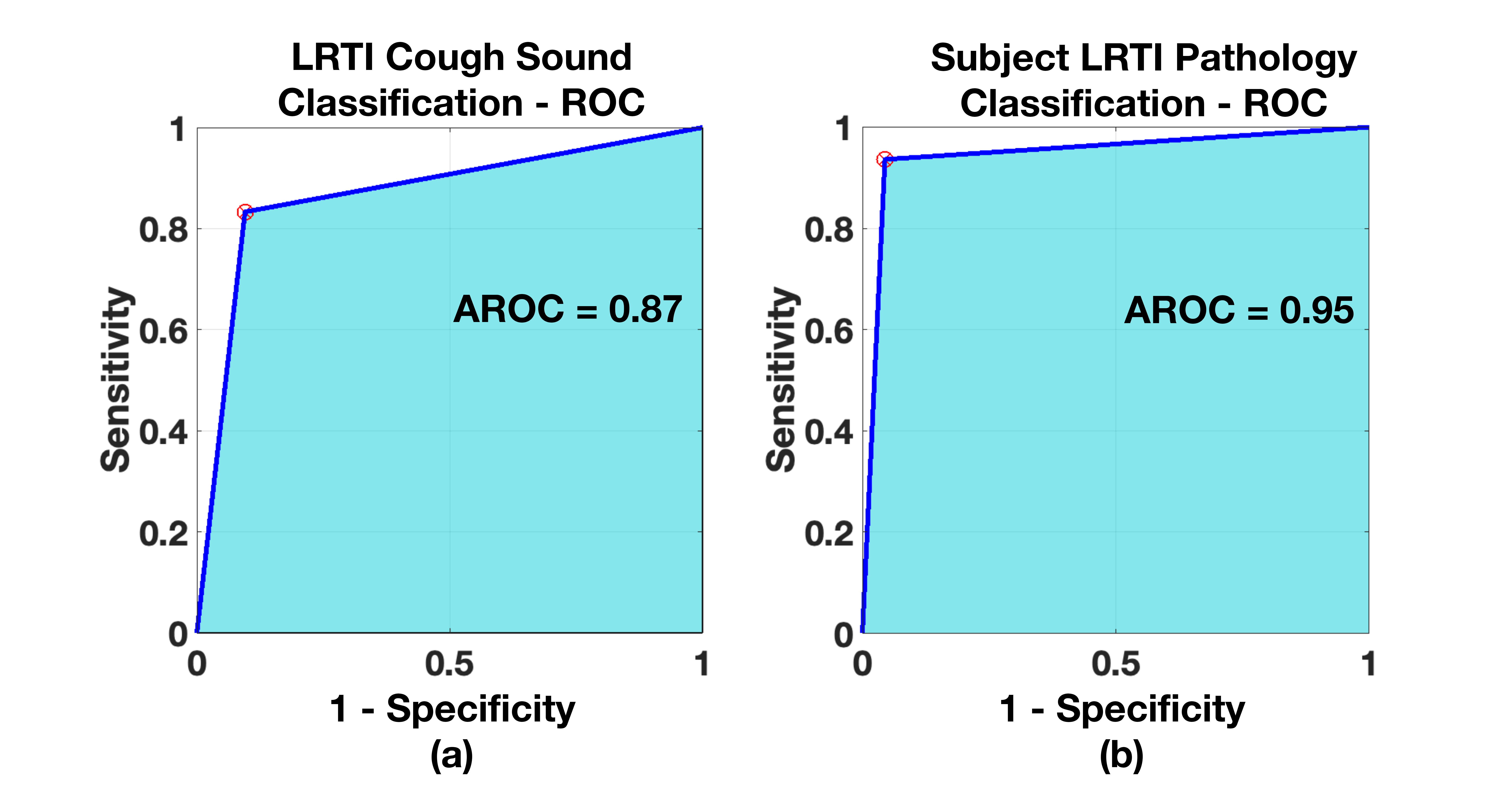}
 \caption{ROC - \textit{Healthy vs LRTI Model} (a) when classifying LRTI coughs (b) when classifying subject for LRTI.}
 \label{fig:LRTIresult_AROC}
\end{figure}

\begin{figure}[H]
 \centering
 \includegraphics[width=1\linewidth]{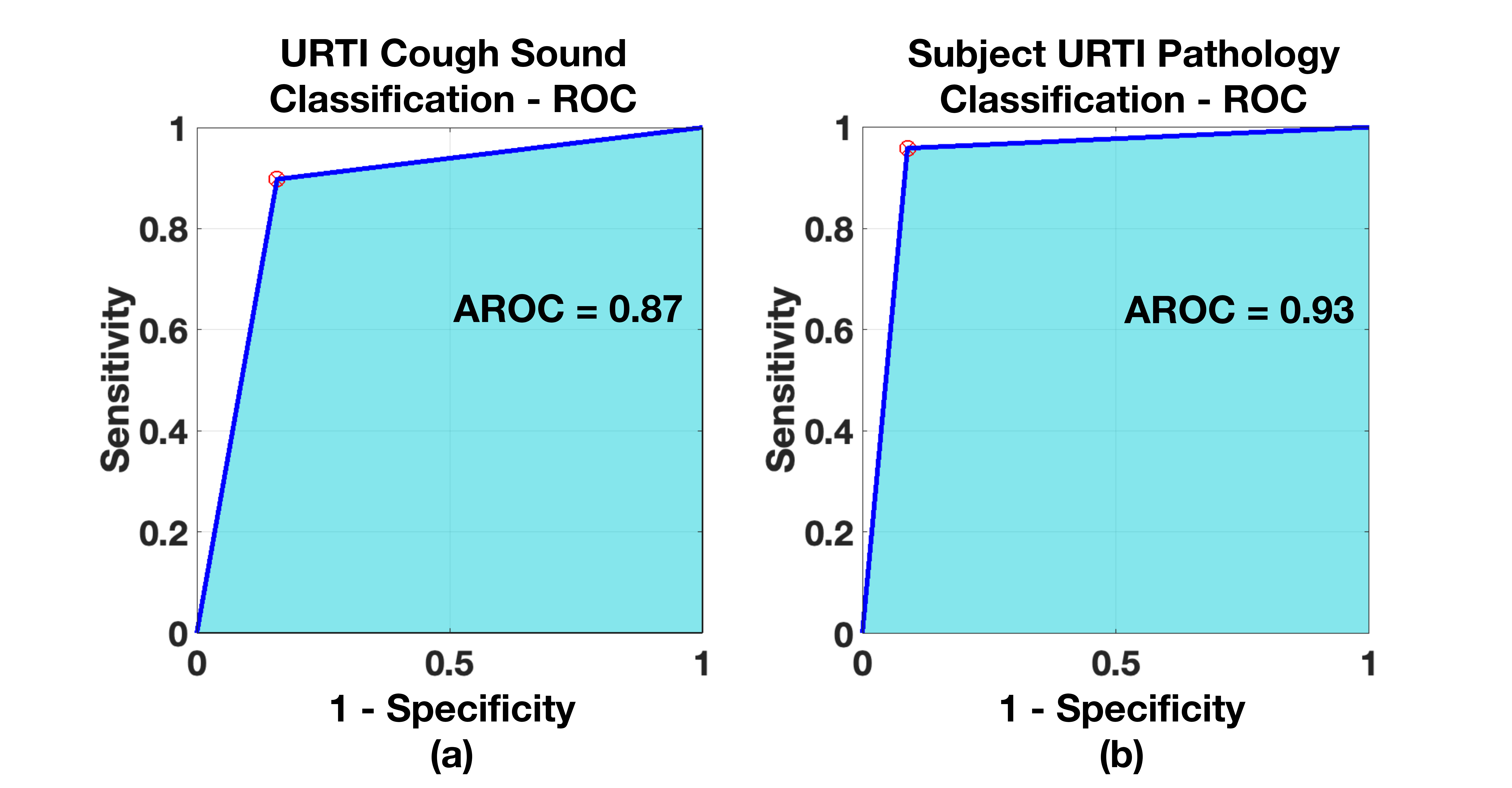}
 \caption{ROC - \textit{Healthy vs URTI Model} (a) when classifying URTI coughs (b) when classifying subject for URTI.
}
 \label{fig:URTIresult_AROC}
\end{figure}

\begin{figure}[H]
 \centering
  \includegraphics[width=1\linewidth]{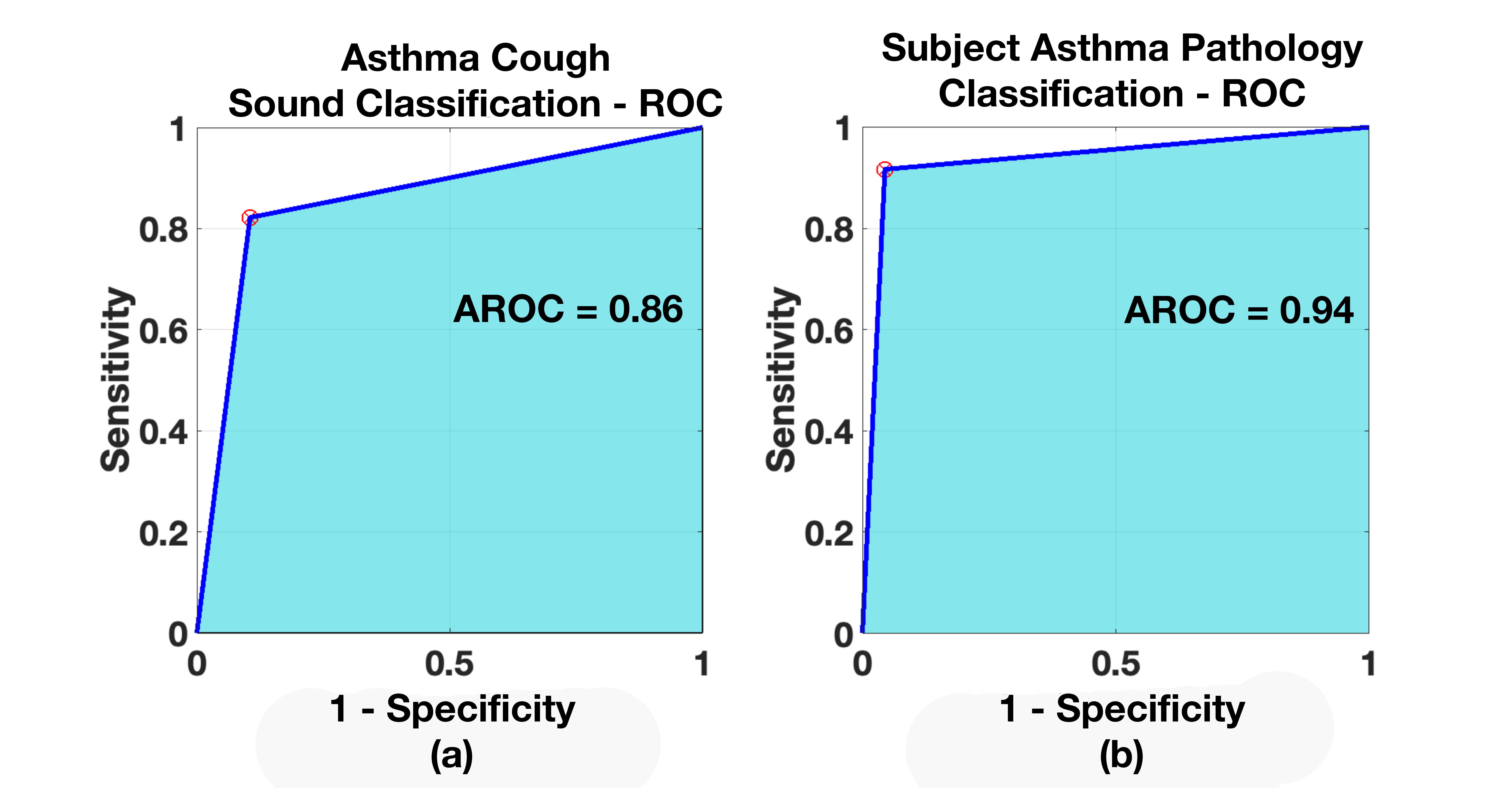}
 \caption{ROC - \textit{Healthy vs Asthma Model} (a) when classifying Asthmatic coughs (b) when classifying subject for Asthma.}
 \label{fig:Asthmaresult_AROC}
\end{figure}

Receiver operating characteristics were created for all three models, both for the case of cough and pathology classification. The ROCs are shown in Figure \ref{fig:LRTIresult_AROC}, \ref{fig:URTIresult_AROC}, \ref{fig:Asthmaresult_AROC} and the resulting AROC is shown in Table \ref{tab:classAROC}. The AROC values are convincingly higher for all the pathology screening results (exceeding 93\%) compared to the individual cough classification models. They support the finding from Table \ref{tab:classaccura_pathoone} and the corresponding confusion matrices.

\begin{table}
\small
\centering
\caption{Area under the receiver operating curve (AROC).}
\label{tab:classAROC}
\begin{tabular}{l|cc}
\hline
Model & \begin{tabular}[c]{@{}c@{}}Individual Cough \\ Classification \\ AROC  \end{tabular} & \begin{tabular}[c]{@{}c@{}}Respiratory \\Pathology \\ Classification of subject \\ AROC \\ based on\\  entire cough epoch \end{tabular} \\

\hline
\textit{Healthy vs LRTI Model} & 0.87 & 0.95 \\
\textit{Healthy vs URTI Model} & 0.87 & 0.93  \\
\textit{Healthy vs Asthma Model} & 0.86 & 0.94 \\
\hline
\end{tabular}
\vspace{-2mm}
\end{table}

\subsubsection{\textit{Healthy vs Pathology (4-class) Model}}
\begin{figure}
 \centering
 \includegraphics[width=1\linewidth]{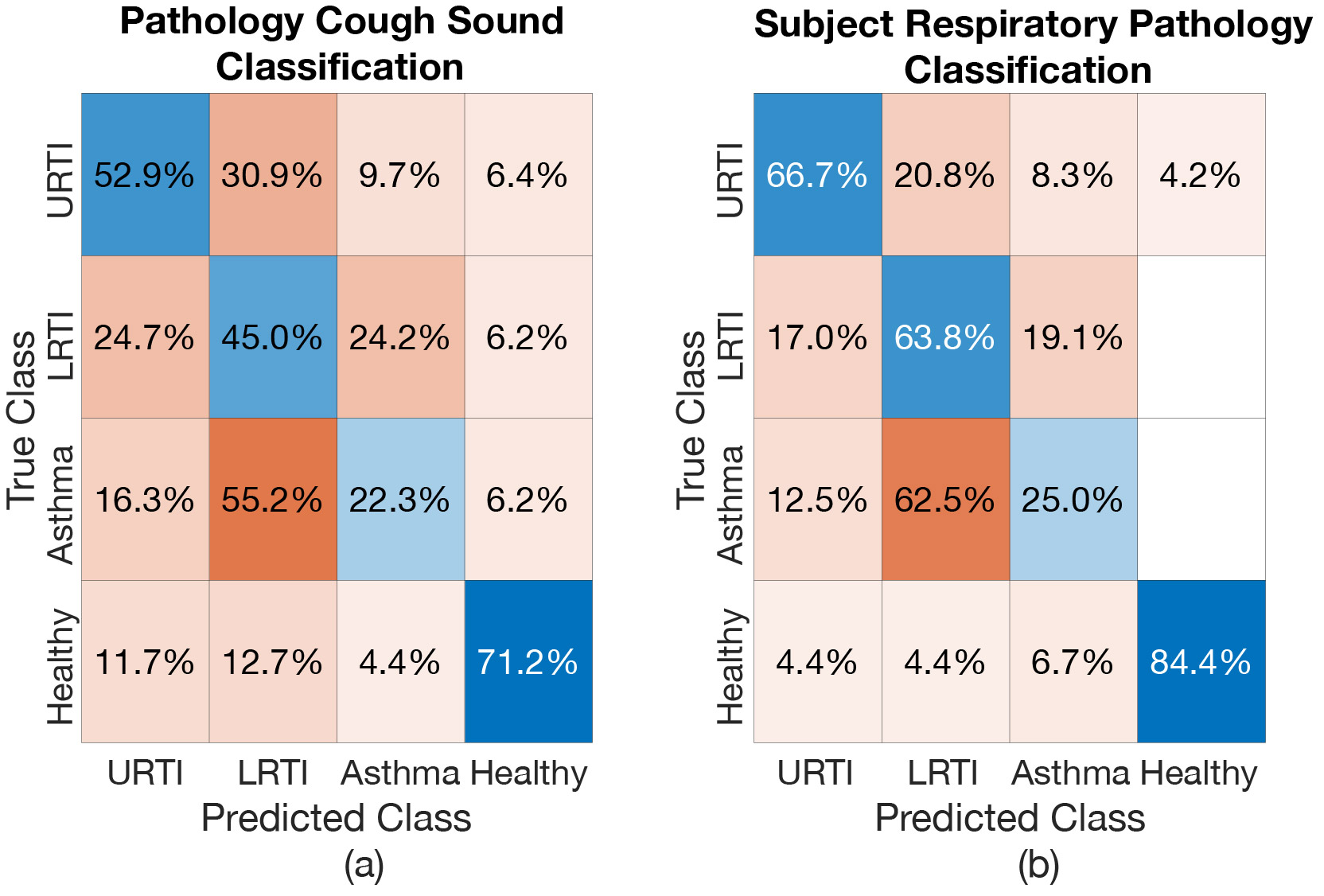}
 \caption{Confusion Matrix - \textit{Healthy vs Pathology (4-class) Model}  (a) when screening coughs (b) when screening for pathology of the subjects.}
 \label{fig:Conf_4class}
\end{figure}

The resulting performance of the proposed model when trained to classify different respiratory pathological coughs and healthy-voluntary coughs (i.e. four classes model) is shown in Table \ref{Tab:Cough_path_4class}. The subject respiratory pathology classification result for this four-class classification, based on the most frequent (mode) prediction outcome for all cough epochs of a subject, is shown in Table \ref{Tab:Pathology_path_4class}. The overall classification accuracy of both cough classification and each pathology classification is lower compared to the results shown in Table \ref{tab:classaccura_model1} and \ref{tab:classaccura_pathoone}. The classification accuracy for the healthy-voluntary cough class and the subsequent respiratory pathology classification is relatively high (71.2\% and 84.4\%, respectively). However, the classification accuracy of pathological cough classes is relatively low. The Asthma class has the highest misclassification rate among the three investigated respiratory conditions. The confusion matrices are shown to further understand this classification result (See Figure \ref{fig:Conf_4class} (a)). It is interesting to note in the respiratory pathology classification results ((See Figure \ref{fig:Conf_4class} (b))) that none of the subjects with LRTI and Asthma are misclassified as healthy and only one subject having URTI is misclassified as healthy (4.2\% out of 24 subject with URTI tested will be one subject). However, seven healthy subjects were misclassified to have some kind of respiratory problems (of these seven, two were misclassified as having URTI, another two were misclassified as having LRTI and another three misclassified as having Asthma). Among the three respiratory conditions, as mentioned earlier, Asthma was the most misclassified pathology (15 subjects out of 24 with asthma were misclassified as having LRTI). Even though there is high misclassification rate among the three investigated respiratory conditions, in summary, this four-class classification model has a classification accuracy of 84.5\% for correctly identifying healthy subjects and 95.8\% accuracy for identifying subjects with respiratory issues , see Table \ref{tab:classaccura_modelcombined_4}. 

\begin{table}[H]
\centering
\caption{Cough classification accuracy of \textit{ Healthy vs Pathology 4 class Model}.}
\label{tab:coughdataset}
\begin{tabular}{ccccc}
\toprule
\vspace{.1cm}
 Overall & Healthy  & Asthma & URTI & LRTI \\
 cough  & cough  & cough & cough & cough  \\
  classification  & classification  & classification & classification & classification  \\
 accuracy & accuracy & accuracy & accuracy & accuracy \\
  (in \%) & (in \%) & (in \%) & (in \%) & (in \%) \\
\midrule

47.9     & 71.2 & 22.3&52.9&45.0  \\

\bottomrule
\end{tabular}
\label{Tab:Cough_path_4class}
\end{table}

\begin{table}
\centering
\caption{Pathology classification accuracy for \textit{Healthy vs Pathology 4 class Model}.}
\label{tab:coughdataset}
\begin{tabular}{ccccc}
\toprule
\vspace{.1cm}
 \begin{tabular}[c]{@{}c@{}}Overall\\ Respiratory \\Pathology \\ Classification \\ of subject \\ accuracy \\ based on\\  entire \\cough epoch\\(in \%) \end{tabular}  & \begin{tabular}[c]{@{}c@{}} Healthy \\Subject \\ Classification \\ accuracy \\ based on\\  entire \\cough epoch\\(in \%) \end{tabular}   & \begin{tabular}[c]{@{}c@{}}Asthmatic \\subject \\Classification \\ accuracy \\ based on\\  entire \\cough epoch\\(in \%) \end{tabular}  & \begin{tabular}[c]{@{}c@{}}URTI \\subject \\Classification \\ accuracy \\ based on\\  entire \\cough epoch\\(in \%) \end{tabular}  & \begin{tabular}[c]{@{}c@{}}LRTI \\subject \\Classification \\ accuracy \\ based on\\  entire \\ cough epoch\\(in \%) \end{tabular}  \\
\midrule

60.0     & 84.5 & 25.0&66.7&63.8  \\

\bottomrule
\end{tabular}
\label{Tab:Pathology_path_4class}
\end{table}

\begin{table}
\small
\centering
\caption{ Accuracy of\textit{Healthy vs Pathology 4 class Model}}
\label{tab:classaccura_modelcombined_4}
\begin{tabular}{l|cc}
\toprule
Model & \begin{tabular}[c]{@{}c@{}}Healthy \\ Classified \\ as healthy \\(in \%) \end{tabular} & \begin{tabular}[c]{@{}c@{}}Subjects having \\ respiratory conditions \\ and classified to have \\some kind of respiratory conditions\\(in \%) \end{tabular} \\
\midrule
\textit{Healthy vs pathology Model} & 84.5 & 95.8 \\
\\
\hline

\end{tabular}
\end{table}



\section{Conclusion}

A classifier was developed based on a BiLSTM model trained using Mel Frequency Cepstral Coefficient features  that can differentiate cough sounds from healthy children with no active respiratory pathology to those with active pathological respiratory conditions such as Asthma, URTI and LRTI. Four classifiers were trained as part of this investigation. The resulting trained model that classifies  cough sounds into healthy/pathological in general or healthy/belonging to LRTI, URTI and Asthma resulted in classification accuracy exceeding 84\% when predicting a clinician's diagnosis. When a respiratory pathology classification of subject was performed using the mode of the prediction results across the multiple cough epochs from a particular subject, the resulting classification accuracy exceeded 91\%. The classification accuracy of the model was compromised when trained to classify all the four classes of cough categories in one-shot. However, most of the misclassification were happened within the pathological classes where one class of pathological cough is often misclassified into having another pathology. If one ignores such misclassification and considers healthy cough to be that from a healthy subject and pathological cough to have come from subject with some kind of pathology, then the overall accuracy of the classifier is above 84\%. This is a first step towards developing a highly efficient deep neural network model that can differentiate between different pathological cough sounds. Such a model could support physicians in creating a differential screening of respiratory conditions that present with cough, and will thus add value to health status monitoring and triaging in medical care.


\vspace{6pt}



\authorcontributions{For research articles with several authors, a short paragraph specifying their individual contributions must be provided. The following statements should be used ``Conceptualization, X.X. and Y.Y.; methodology, X.X.; software, X.X.; validation, X.X., Y.Y. and Z.Z.; formal analysis, X.X.; investigation, X.X.; resources, X.X.; data curation, X.X.; writing---original draft preparation, X.X.; writing---review and editing, X.X.; visualization, X.X.; supervision, X.X.; project administration, X.X.; funding acquisition, Y.Y. All authors have read and agreed to the published version of the manuscript.'', please turn to the \href{http://img.mdpi.org/data/contributor-role-instruction.pdf}{CRediT taxonomy} for the term explanation. Authorship must be limited to those who have contributed substantially to the work~reported.}

\funding{Please add: ``This research received no external funding'' or ``This research was funded by NAME OF FUNDER grant number XXX.'' and and ``The APC was funded by XXX''. Check carefully that the details given are accurate and use the standard spelling of funding agency names at \url{https://search.crossref.org/funding}, any errors may affect your future funding.}

\institutionalreview{The study was conducted under Singhealth IRB no. 2016/2416 and ClinialTrials.gov no. NCT03169699 and funded by SMART no. ING000091-ICT.}


\informedconsent{Any research article describing a study involving humans should contain this statement. Please add ``Informed consent was obtained from all subjects involved in the study.'' OR ``Patient consent was waived due to REASON (please provide a detailed justification).'' OR ``Not applicable'' for studies not involving humans. You might also choose to exclude this statement if the study did not involve humans.

Written informed consent for publication must be obtained from participating patients who can be identified (including by the patients themselves). Please state ``Written informed consent has been obtained from the patient(s) to publish this paper'' if applicable.}

\dataavailability{Please contact authors to access the data used in this study}


\acknowledgments{
We thank Ariv K (from SUTD for helping with audio segmentation), Teng SS, Dianna Sri Dewi and Foo Chuan Ping (from KK Women's and Children's Hospital, Singapore for coordinating the recruitment of patients and research project administration).}

\conflictsofinterest{Declare conflicts of interest or state ``The authors declare no conflict of interest.'' Authors must identify and declare any personal circumstances or interest that may be perceived as inappropriately influencing the representation or interpretation of reported research results. Any role of the funders in the design of the study; in the collection, analyses or interpretation of data; in the writing of the manuscript, or in the decision to publish the results must be declared in this section. If there is no role, please state ``The funders had no role in the design of the study; in the collection, analyses, or interpretation of data; in the writing of the manuscript, or in the decision to publish the~results''.}

\end{paracol}
\reftitle{References}



\bibliography{mybib}

\end{document}